\documentclass[10pt,twocolumn,letterpaper]{article}

\usepackage[pagenumbers]{cvpr} % arXiv version with page numbers

\usepackage{amsmath,amssymb}
\usepackage{booktabs}
\usepackage{multirow}
\usepackage[table]{xcolor}
\usepackage{graphicx}
\usepackage{xspace}
\usepackage{microtype}

\newcommand{\method}{GazeQwen\xspace}
\newcommand{\benchmark}{StreamGaze\xspace}

\definecolor{pastColor}{HTML}{E7F2DC}
\definecolor{presentColor}{HTML}{FBF5DC}
\definecolor{proactiveColor}{HTML}{FBE4E7}

\definecolor{cvprblue}{rgb}{0.21,0.49,0.74}
\usepackage[pagebackref,breaklinks,colorlinks,allcolors=cvprblue]{hyperref}
\usepackage[capitalize]{cleveref}

\def\paperID{*****}
\def\confName{CVPR}
\def\confYear{2026}

\title{GazeQwen: Lightweight Gaze-Conditioned LLM Modulation\\for Streaming Video Understanding}

\author{
Trong-Thang Pham$^{1}$\thanks{Corresponding author.} \quad
Hien Nguyen$^{2}$ \quad
Ngan Le$^{1}$\\[4pt]
$^{1}$University of Arkansas \quad
$^{2}$University of Houston\\[2pt]
{\tt\small tp030@uark.edu \quad hvnguy35@central.uh.edu \quad thile@uark.edu}
}

\begin{document}
\maketitle

\begin{abstract}
Current multimodal large language models (MLLMs) cannot effectively utilize eye-gaze information for video understanding, even when gaze cues are supplied via visual overlays or text descriptions.
We introduce \method{}, a parameter-efficient approach that equips an open-source MLLM with gaze awareness through hidden-state modulation.
At its core is a compact gaze resampler (${\sim}$1--5\,M trainable parameters) that encodes V-JEPA~2.1 video features together with fixation-derived positional encodings and produces additive residuals injected into selected LLM decoder layers via forward hooks.
An optional second training stage adds low-rank adapters (LoRA) to the LLM for tighter integration.
Evaluated on all 10 tasks of the \benchmark{} benchmark, \method{} reaches 63.9\% accuracy, a +16.1 point gain over the same Qwen2.5-VL-7B backbone with gaze as visual prompts and +10.5 points over GPT-4o, the highest score among all open-source and proprietary models tested.
These results suggest that learning \emph{where} to inject gaze within an LLM is more effective than scaling model size or engineering better prompts. All code and checkpoints are available at \url{https://github.com/phamtrongthang123/gazeqwen}.
\end{abstract}

\section{Introduction}

Applications such as AR-glass assistants and embodied agents require models that interpret continuous video streams in real time~\cite{lin2024streamingbench,chen2024videollm,fu2025vispeak}.
In egocentric settings, eye gaze reveals the user's focus of attention and likely next actions~\cite{gazean1,gazean2,egtea}, yet how to make MLLMs actually \emph{use} gaze data remains open.

The \benchmark{} benchmark~\cite{streamgaze2025} quantifies this gap: 8{,}521 gaze-conditioned QA pairs across 285 egocentric videos.
All tested MLLMs, including GPT-4o (53.4\%) and Qwen2.5-VL-7B (47.8\%), lag well behind humans (82.7\%).
Notably, none of the three prompting strategies tested in that work (textual coordinates, visual overlays, saliency maps) reliably surpass the gaze-free baseline, suggesting that surface-level formatting is insufficient.
Prior gaze-aware approaches either overlay gaze markers without training signal~\cite{peng2025eye}, or train task-specific models that do not generalize to streaming scenarios~\cite{ilaslan2023gazevqa}.

We take a different approach with \method{}.
Instead of modifying the input or retraining the backbone, we attach a small resampler module that \emph{directly steers the LLM's internal representations} based on gaze.
It encodes fixation coordinates as sinusoidal positional signals, fuses them with frozen V-JEPA~2.1 video features~\cite{bardes2024revisiting}, and injects the result as additive residuals at selected LLM layers via PyTorch forward hooks.
An optional second stage adds LoRA adapters~\cite{hu2022lora} for tighter integration.
The design is \textbf{low-cost} (${\sim}$5--9\,M trainable parameters, single-GPU training), \textbf{modular} (gaze module attaches/detaches without reloading weights), and \textbf{effective}: \method{} scores 63.9\% on \benchmark{}, +16.1pp over the same Qwen2.5-VL backbone and +10.5pp over GPT-4o.

\section{Method}

\begin{figure}[t]
    \centering
    \includegraphics[width=\linewidth]{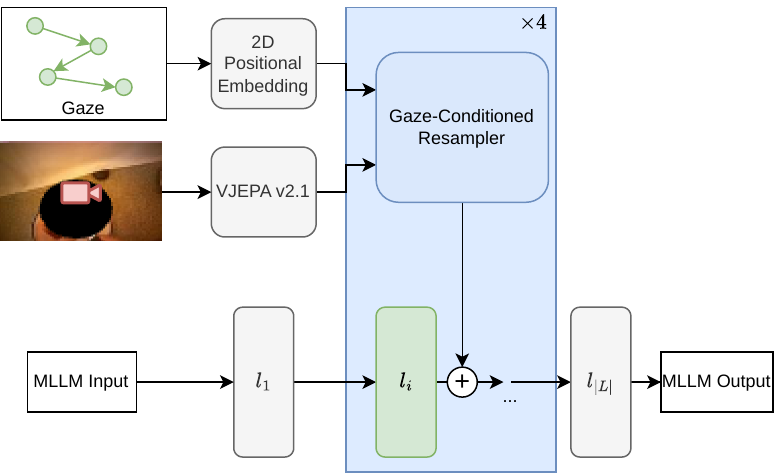}
    \caption{\textbf{\method{} overview.} A small gaze resampler (${\sim}$1--5\,M parameters) encodes V-JEPA~2.1 video features together with fixation-derived positional encodings and produces additive residuals injected into four evenly-spaced LLM decoder layers via forward hooks.}
    \label{fig:architecture}
\end{figure}

\method{} equips an open-source MLLM with gaze awareness through hidden-state modulation (\Cref{fig:architecture}).
A compact gaze resampler encodes V-JEPA~2.1 video features together with fixation-derived positional encodings and produces additive residuals injected into selected LLM decoder layers via forward hooks.

\subsection{Inputs}

\paragraph{Video features.}
Given a video clip, we sample $T$ frames and pass them through a frozen V-JEPA~2.1 encoder~\cite{bardes2024revisiting} (ViT-B/16, $384{\times}384$).
We use a \emph{separate} visual encoder rather than reusing the MLLM's own vision backbone because the host model's visual features are optimized for language grounding and may discard fine-grained spatial detail needed for gaze alignment; V-JEPA's self-supervised spatiotemporal representations retain this information without task-specific bias.
For each temporal step $t \in \{1, \dots, T\}$, the encoder produces a spatial feature grid $\mathbf{F}_t \in \mathbb{R}^{HW \times d_v}$, where $H, W$ are the spatial grid dimensions and $d_v{=}768$.
The V-JEPA output is interpolated (trilinear temporal, bilinear spatial) to align with the host LLM's visual token grid.

\paragraph{Gaze scanpath.}
Eye gaze is represented as a scanpath $\mathcal{S} = \{(x_i, y_i, t_i, \Delta t_i)\}_{i=1}^{N}$, where $(x_i, y_i) \in [0,1]^2$ are normalized fixation coordinates, $t_i$ is the midpoint timestamp, and $\Delta t_i$ is the duration.
A fixation is \emph{active} at frame time $t$ if $|t - t_i| \leq \Delta t_i / 2$; we denote the active set as $\mathcal{A}_t$.

\paragraph{Gaze encoding.}
Active fixations are encoded using DETR-style sinusoidal positional encoding~\cite{carion2020end} with latent dimension $d_l{=}256$.
Sinusoidal encodings are parameter-free and provide a smooth, continuous mapping from coordinates to high-dimensional space, making them well-suited for the sparse and variable-count nature of fixation inputs.
Multiple active fixations are averaged:
\begin{equation}
    \mathbf{g}_t = \frac{1}{|\mathcal{A}_t|}\sum_{i \in \mathcal{A}_t} \text{PE}(x_i, y_i) \;\in\; \mathbb{R}^{d_l}.
\end{equation}
When $\mathcal{A}_t = \emptyset$, we set $\mathbf{g}_t = \mathbf{0}$.

\subsection{Gaze-Conditioned Resampler}

The resampler takes $\mathbf{F}_t$ and $\mathbf{g}_t$ and produces a residual in the LLM's hidden-state space.
V-JEPA tokens are projected to $\mathbf{X} = \mathbf{F}_t \mathbf{W}_{\text{in}} \in \mathbb{R}^{HW \times d_l}$.
The gaze vector is broadcast to $\mathbf{G} \in \mathbb{R}^{HW \times d_l}$.
$K{=}32$ learnable latents $\mathbf{L} \in \mathbb{R}^{K \times d_l}$ serve as an information bottleneck: by forcing all visual--gaze interactions through a small set of latent vectors, the resampler must learn to distill the scene into a compact, gaze-relevant summary rather than simply copying all spatial tokens.

Over $B{=}2$ cross-attention blocks, the latents attend over $[\mathbf{X}; \mathbf{L}]$ with gaze-biased keys.
Adding the gaze encoding to the \emph{keys} rather than the queries or values steers the attention distribution: tokens near the fixation point receive higher attention weight, while the aggregated content (values) remains unaltered, preserving visual semantics while spatially re-weighting them:
\begin{align}
    \mathbf{Q} &= \mathbf{L}\mathbf{W}_Q, \quad
    \mathbf{K} = [\mathbf{X}; \mathbf{L}]\mathbf{W}_K + [\mathbf{G}; \mathbf{0}]\mathbf{W}_G, \\
    \mathbf{V} &= [\mathbf{X}; \mathbf{L}]\mathbf{W}_V, \\
    \mathbf{L} &\leftarrow \mathbf{L} + \text{softmax}\!\left(\tfrac{\mathbf{Q}\mathbf{K}^\top}{\sqrt{d_l}}\right)\mathbf{V},
\end{align}
followed by LayerNorm + FFN ($4 d_l$ hidden).
After the blocks, spatial tokens read from the updated latents via reverse cross-attention.
This reverse step is necessary because the forward pass compresses $HW$ spatial tokens into only $K{=}32$ latents; the reverse cross-attention expands the gaze-modulated representation back to the original spatial resolution so that each visual token in the LLM receives a position-specific residual.
A zero-initialized output projection maps to $d_{\text{llm}}{=}3{,}584$, ensuring the module has no effect at the start of training and the pretrained LLM representations are not disrupted:
\begin{equation}
    \mathbf{R}_t = \text{XAttn}(\mathbf{X}, \mathbf{L})\;\mathbf{W}_{\text{out}} \;\in\; \mathbb{R}^{HW \times d_{\text{llm}}}.
\end{equation}

\subsection{Hook-Based LLM Injection}

We inject gaze information at multiple depths of the LLM rather than at a single point, because different decoder layers encode progressively more abstract representations: early layers capture low-level spatial layout while later layers handle high-level semantics. Injecting at four evenly-spaced layers ($l \in \{6, 13, 20, 27\}$ out of 28) via PyTorch forward hooks lets gaze influence all levels of abstraction without modifying the model's forward pass code:
\begin{equation}
    \mathbf{h}_l^{(v)} \leftarrow \mathbf{h}_l^{(v)} + \alpha_l \cdot f_\theta^{(l)}(\mathbf{F}_t, \mathbf{g}_t),
\end{equation}
where $\alpha_l$ is a learned per-layer amplitude scalar that lets the model control how strongly gaze modulates each depth.
Each layer has an \emph{independent} resampler so that the gaze signal can be tailored to the representation space at each depth; a shared resampler would be forced to produce a single residual that is simultaneously appropriate for all layers.
Additional hooks on the visual encoder and model input locate the visual-token positions in the LLM sequence.
Text tokens are not modified.

\subsection{Training}

\paragraph{Two-stage training.}
Training proceeds in two stages to decouple gaze alignment from LLM adaptation.
\emph{Stage~1:} Only the four resamplers (${\sim}$1--5\,M params) are trained with 4-way cross-entropy over answer-token logits (A/B/C/D).
This stage teaches the resamplers to produce useful gaze residuals while the LLM remains fully frozen, preventing catastrophic forgetting.
\emph{Stage~2:} Rank-8 LoRA adapters~\cite{hu2022lora} ($\alpha{=}16$, ${\sim}$3.5\,M params) on LLM Q/V projections are added and trained jointly with the resamplers, allowing the LLM to better incorporate the gaze signal by adjusting its own attention patterns.
Both stages use AdamW (lr~$3{\times}10^{-4}$, weight decay~$10^{-2}$), 20 warmup steps, gradient accumulation of 8, and up to 20 epochs.
Data is split 70/15/15 by video to prevent leakage.

\section{Experiments}

\subsection{Setup}

\paragraph{Benchmark.}
We evaluate on \benchmark{}~\cite{streamgaze2025}, a gaze-conditioned video QA benchmark containing 8{,}521 multiple-choice questions (4 options each) derived from 285 egocentric videos with eye-tracking data.
The benchmark defines 10 task types grouped into three temporal categories~\cite{streamgaze2025}:
\emph{Past} tasks (NFI, OTP, SR, GSM) require reasoning over the full viewing history;
\emph{Present} tasks (OI-Easy, OI-Hard, OAR, FAP) operate within a 60-second context window around the query time;
\emph{Proactive} tasks (GTA, OAA) require anticipating future events or alerting the user based on current gaze patterns.
We use a 70/15/15 video-level split, yielding 1{,}055 test QA pairs.

\paragraph{Backbone.}
Our base MLLM is Qwen2.5-VL-7B-Instruct~\cite{bai2025qwen25}, loaded with eager attention to allow PyTorch forward hook registration on individual decoder layers.
Visual features come from a frozen V-JEPA~2.1 encoder (ViT-B/16, $384{\times}384$ input resolution)~\cite{bardes2024revisiting}, whose spatiotemporal features are interpolated (trilinear temporal, bilinear spatial) to match Qwen's visual token grid.
Both the MLLM and V-JEPA backbones remain frozen throughout training; only the resampler modules and optional LoRA adapters are updated.

\paragraph{Baselines.}
We compare against four categories of models.
\emph{Closed-source MLLMs:} GPT-4o~\cite{openai2024gpt4technicalreport} and Claude Sonnet/Opus~4~\cite{anthropic2025claude4}, evaluated with gaze supplied as textual coordinate prompts.
\emph{Open-source MLLMs:} Qwen2.5-VL~\cite{bai2025qwen25}, InternVL3.5~\cite{wang2025internvl3}, VITA~1.5~\cite{fu2025vita}, MiniCPM-V~\cite{yao2024minicpm}, and Kangaroo~\cite{liu2024kangaroo}.
\emph{Streaming MLLMs:} ViSpeak~\cite{fu2025vispeak}, Dispider~\cite{qian2025dispider}, Flash-VStream~\cite{zhang2024flashvstream}, and VideoLLM-online~\cite{chen2024videollm}.
\emph{Gaze-specialized:} AssistGaze~\cite{ilaslan2023gazevqa}, a fine-tuned model trained on GazeQA data.
All baseline numbers are taken from \benchmark{}~\cite{streamgaze2025}.

\paragraph{Evaluation protocol.}
Each QA pair is evaluated independently (batch size 1) to accommodate variable video lengths.
The predicted answer is the option (A/B/C/D) with the highest logit at the last token position.
We report per-task accuracy and the unweighted mean across all 10 tasks.

\subsection{Main Results}

\Cref{tab:main_results} summarizes accuracy on each task.
\method{} reaches \textbf{63.9\%} overall, the highest among all systems including proprietary ones:
\begin{itemize}
    \item \textbf{+10.5pp} over GPT-4o (53.4\%), despite using a 7B open-weight model versus a much larger proprietary system with gaze supplied as text coordinates.
    \item \textbf{+16.1pp} over the same Qwen2.5-VL backbone with gaze supplied as visual prompts (47.8\%), demonstrating that internal hidden-state modulation is far more effective than surface-level input formatting.
\end{itemize}

\paragraph{Where does gaze help most?}
The per-task breakdown reveals that improvements concentrate on tasks where spatial gaze information directly reduces ambiguity.
\textbf{OAA} (Object Appearance Alert) gains +34.1pp, as proactive alerts about object appearances benefit directly from knowing which objects the user is currently attending to.
\textbf{OTP} (Object Tracking in Past) gains +29.4pp, as the accumulated fixation history provides a strong temporal signal for tracking attended objects over time.
\textbf{OAR} (Object Attribute Recognition) gains +26.1pp because knowing exactly which object the user fixates narrows the relevant region, making attribute recognition straightforward even when multiple objects with different attributes coexist in the scene.
\textbf{OI-Hard} (Object Identification Hard) gains +20.8pp since gaze disambiguates the target from visually similar nearby objects that confuse the model without spatial guidance.

\paragraph{Where does gaze not help?}
\textbf{OI-Easy} (+0.4pp) and \textbf{FAP} (+1.7pp) show negligible gains, suggesting that easy object identification is already achievable without gaze disambiguation, and predicting future actions requires higher-level temporal reasoning beyond where the user is currently looking.
\textbf{SR} (Scene Recall, +9.8pp) shows a moderate gain; the task asks about previously seen background context, where gaze localization is less critical since the answer depends on global scene memory rather than fixation targets.

\begin{table*}[t]
  \centering
  \footnotesize
  \caption{\textbf{Results on the \benchmark{} benchmark.}
  We report accuracy on all 10 task types.
  Baseline numbers are from~\cite{streamgaze2025}; \method{} is our evaluation on the Qwen2.5-VL-7B backbone.
  Overall is the mean over the 10 tasks shown.
  Best per-task (excluding Human) is \textbf{bolded}.
  }
  \vspace{-0.05in}
  \resizebox{\linewidth}{!}{
  \begin{tabular}{lcc|cccc|cccc|cc|c}
    \toprule
    \multirow{2}{*}{Method} &
    \multirow{2}{*}{Params} &
    \multirow{2}{*}{Frames} &
    \multicolumn{4}{c|}{\cellcolor{pastColor}\textbf{Past}} &
    \multicolumn{4}{c|}{\cellcolor{presentColor}\textbf{Present}} &
    \multicolumn{2}{c|}{\cellcolor{proactiveColor}\textbf{Proactive}} &
    \multirow{2}{*}{\textbf{Overall}} \\

    & & &
    \cellcolor{pastColor}NFI &
    \cellcolor{pastColor}OTP &
    \cellcolor{pastColor}SR &
    \cellcolor{pastColor}GSM &
    \cellcolor{presentColor}OI (E) &
    \cellcolor{presentColor}OI (H) &
    \cellcolor{presentColor}OAR &
    \cellcolor{presentColor}FAP &
    \cellcolor{proactiveColor}GTA &
    \cellcolor{proactiveColor}OAA &
    \\

    \midrule
    Human & -- & -- & 0.700 & 0.889 & 0.903 & 0.707 & 0.960 & 0.920 & 0.800 & 0.840 & 0.765 & 0.780 & 0.827 \\

    \midrule
    \rowcolor{gray!15} \multicolumn{14}{c}{\textit{Closed-Source MLLMs}}\\
    \midrule
    GPT-4o~\cite{openai2024gpt4technicalreport} & -- & 16 &
    0.601 & 0.449 & 0.535 & 0.580 &
    \textbf{0.729} & 0.730 & 0.596 & 0.370 &
    0.597 & 0.149 & 0.534 \\
    Claude Sonnet4~\cite{anthropic2025claude4} & -- & 16 &
    0.500 & 0.554 & 0.425 & 0.325 &
    0.521 & 0.533 & 0.561 & \textbf{0.439} &
    0.535 & 0.350 & 0.474 \\
    Claude Opus4~\cite{anthropic2025claude4} & -- & 16 &
    0.372 & 0.392 & 0.460 & 0.166 &
    0.431 & 0.436 & 0.430 & 0.351 &
    0.466 & 0.490 & 0.399 \\

    \midrule
    \rowcolor{gray!15} \multicolumn{14}{c}{\textit{GazeQA-based Fine-tuned Models}}\\
    \midrule
    AssistGaze~\cite{ilaslan2023gazevqa} & 26M & 32 &
    0.294 & 0.131 & 0.310 & 0.294 &
    0.278 & 0.250 & 0.109 & 0.254 &
    N/A & N/A & 0.240 \\

    \midrule
    \rowcolor{gray!15} \multicolumn{14}{c}{\textit{Open-Source MLLMs}}\\
    \midrule
    Qwen2.5-VL~\cite{bai2025qwen25} & 7B & Adapt. &
    0.518 & 0.350 & 0.450 & 0.483 &
    0.590 & 0.558 & 0.548 & 0.391 &
    0.486 & 0.407 & 0.478 \\
    InternVL3.5~\cite{wang2025internvl3} & 8B & Adapt. &
    0.490 & 0.311 & \textbf{0.573} & 0.548 &
    0.627 & 0.628 & 0.466 & 0.372 &
    0.373 & 0.051 & 0.444 \\
    VITA 1.5~\cite{fu2025vita} & 7B & 16 &
    0.474 & 0.365 & 0.346 & 0.378 &
    0.455 & 0.396 & 0.437 & 0.370 &
    0.351 & 0.267 & 0.384 \\
    MiniCPM-V~\cite{yao2024minicpm} & 8B & 32 &
    0.430 & 0.374 & 0.354 & 0.296 &
    0.334 & 0.379 & 0.438 & 0.345 &
    0.480 & 0.216 & 0.365 \\
    Kangaroo~\cite{liu2024kangaroo} & 7B & 64 &
    0.363 & 0.365 & 0.319 & 0.275 &
    0.454 & 0.484 & 0.402 & 0.412 &
    0.242 & 0.198 & 0.351 \\

    \midrule
    \rowcolor{gray!15} \multicolumn{14}{c}{\textit{Open-Source Streaming MLLMs}}\\
    \midrule
    ViSpeak~\cite{fu2025vispeak} & 7B & 1\,fps &
    0.463 & 0.358 & 0.417 & 0.473 &
    0.572 & 0.581 & 0.406 & 0.309 &
    \textbf{0.635} & 0.458 & 0.467 \\
    Dispider~\cite{qian2025dispider} & 7B & 1\,fps &
    0.366 & 0.365 & 0.381 & 0.263 &
    0.336 & 0.338 & 0.353 & 0.321 &
    0.252 & 0.261 & 0.324 \\
    Flash-VStream~\cite{zhang2024flashvstream} & 7B & 1\,fps &
    0.249 & 0.202 & 0.336 & 0.220 &
    0.289 & 0.147 & 0.044 & 0.280 &
    0.443 & 0.217 & 0.243 \\
    VideoLLM-online~\cite{chen2024videollm} & 8B & 2\,fps &
    0.000 & 0.000 & 0.000 & 0.000 &
    0.006 & 0.006 & 0.002 & 0.000 &
    0.458 & 0.333 & 0.081 \\

    \midrule
    \rowcolor{gray!15} \multicolumn{14}{c}{\textit{Ours}}\\
    \midrule
    % Qwen2.5-VL (no gaze) & 7B & Adapt. &
    % 0.363 & 0.373 & 0.548 & 0.435 &
    % 0.539 & 0.497 & 0.504 & 0.314 &
    % 0.625 & 0.497 & 0.470 \\
    \textbf{\method{}} & 7B & Adapt. &
    \textbf{0.657} & \textbf{0.644} & 0.548 & \textbf{0.609} &
    0.594 & \textbf{0.766} & \textbf{0.809} & 0.408 &
    0.603 & \textbf{0.748} & \textbf{0.639} \\
    \rowcolor{green!10}
    \textit{$\Delta$ vs.\ Qwen2.5-VL} & & &
    \textit{+13.9} & \textit{+29.4} & \textit{+9.8} & \textit{+12.6} &
    \textit{+0.4} & \textit{+20.8} & \textit{+26.1} & \textit{+1.7} &
    \textit{+11.7} & \textit{+34.1} & \textit{+16.1} \\

    \bottomrule
  \end{tabular}
  }
  \label{tab:main_results}
  \vspace{-0.1in}
\end{table*}

\vspace{-0.05in}
\subsection{Design Space Analysis}
\vspace{-0.03in}

We swept four design axes during development and report the marginal effect of each choice in \Cref{tab:ablation}.

\begin{table}[t]
  \centering
  \footnotesize
  \caption{\textbf{Ablation over design axes.} Each row shows the best option for that axis and the accuracy spread between best and worst options (measured on the task with largest difference).}
  \vspace{-0.05in}
  \begin{tabular}{llc}
    \toprule
    \textbf{Axis} & \textbf{Best option} & \textbf{Spread} \\
    \midrule
    (G) Gaze encoding    & Coord-PE          & 9.8pp (NFI) \\
    (B) Visual backbone  & V-JEPA 2.1        & 8.1pp (NFI) \\
    (S) Layer sharing    & Per-layer (4 ind.) & 5.2pp (OAR) \\
    (A) LLM adaptation   & LoRA (rank 8)     & 5.2pp (OI-H) \\
    \bottomrule
  \end{tabular}
  \label{tab:ablation}
  \vspace{-0.1in}
\end{table}

\textbf{(G)~Gaze encoding:} Coord-PE (sinusoidal encoding of raw fixation coordinates) outperforms both heatmap-$\tau$ (Voila-style~\cite{voila2024} Gaussian heatmap with learnable temporal decay) and heatmap-dur (duration-weighted Gaussian heatmap).
The key advantage is that Coord-PE preserves sub-pixel spatial precision, whereas heatmap encodings discretize gaze onto the LLM's coarse spatial grid, losing fine-grained positional information.
The largest gap appears on NFI (9.8pp), where precise fixation history is critical for determining which objects were never attended.

\textbf{(B)~Visual backbone:} V-JEPA~2.1 consistently outperforms DINOv2 across matched configurations.
V-JEPA's joint spatiotemporal embeddings (via 3D tubelet patches) capture motion and temporal continuity that DINOv2's frame-independent features miss.
The gap is widest on fixation-sequence tasks (NFI: 8.1pp, OTP: 5.3pp), where temporal context is essential.

\textbf{(S)~Layer sharing:} Per-layer resamplers (four independent modules) outperform a single shared resampler by 5.2pp on OAR.
This confirms that different LLM depths benefit from specialized gaze representations: early layers may need broad spatial modulation while deeper layers require fine-grained object-level focus.

\textbf{(A)~LLM adaptation:} Adding rank-8 LoRA adapters in stage~2 provides a further 5.2pp gain on OI-Hard.
LoRA lets the LLM's own attention patterns adapt to better incorporate the injected gaze residuals, rather than relying solely on additive modulation of frozen representations.

All four axes contribute comparably (5--10pp spread), and the gains are largely complementary: removing any single component degrades performance, confirming that the full combination is needed for the best result.

\section{Conclusion}
\vspace{-0.05in}

\method{} shows that a small resampler (${\sim}$1--5\,M params) with LoRA adapters (${\sim}$3.5\,M params) injecting gaze residuals into a 7B MLLM can outperform much larger systems on gaze-conditioned video QA.
The approach requires no architecture surgery, trains on a single GPU, and the gaze module can be detached for gaze-free operation.
Our findings suggest the bottleneck in gaze-aware video understanding is not model capacity but the absence of a learned pathway from gaze to internal representations.
We release all code to support future work on gaze-guided streaming models.

{
    \small
    \bibliographystyle{ieeenat_fullname}
    \bibliography{main}
}

\end{document}